\documentclass[lettersize, journal]{IEEEtran}
\usepackage[utf8]{inputenc}
\usepackage{amsmath}
\usepackage{empheq}
\usepackage{amsfonts}
\usepackage{bbold}
\usepackage{booktabs}
\usepackage{tikz}
\usetikzlibrary{shapes.geometric}
\usepackage{siunitx}
\usepackage{float}
\usepackage[noadjust]{cite}
\usepackage{hyperref}
\usepackage{graphicx}
\usepackage{subcaption}
\graphicspath{ {./images/} }
\usepackage[english]{babel}
\newtheorem{theorem}{Theorem}

\def\BibTeX{{\rm B\kern-.05em{\sc i\kern-.025em b}\kern-.08em
    T\kern-.1667em\lower.7ex\hbox{E}\kern-.125emX}}
\usepackage{balance}

\DeclareMathOperator*{\argmin}{\arg\!\min}
\DeclareMathSymbol{\shortminus}{\mathbin}{AMSa}{"39}

\makeatletter
\newcommand\footnoteref[1]{\protected@xdef\@thefnmark{\ref{#1}}\@footnotemark}
\makeatother

\title{Integrated Decision Control Approach for Cooperative Safety-Critical Payload Transport in a Cluttered Environment}
\author{Nishanth Rao, Suresh Sundaram, \IEEEmembership{Senior Member, IEEE}
\thanks{Nishanth Rao is with the Artificial Intelligence and Robotics Lab, Department of Aerospace Engineering, Indian Institute of Science, Bengaluru, India, 560012. \href{nishanthrao@iisc.ac.in}{\texttt{nishanthrao@iisc.ac.in}}}
\thanks{Suresh Sundaram is with the Artificial Intelligence and Robotics Lab, Department of Aerospace Engineering, Indian Institute of Science, Bengaluru, India, 560012. \href{vssuresh@iisc.ac.in}{\texttt{vssuresh@iisc.ac.in}}}}

\begin{document}

\maketitle

\begin{abstract}
    In this paper, the problem of coordinated transportation of heavy payload by a team of UAVs in a cluttered environment is addressed. The payload is modeled as a rigid body and is assumed to track a pre-computed global flight trajectory from a start point to a goal point. Due to the presence of local dynamic obstacles in the environment, the UAVs must ensure that there is no collision between the payload and these obstacles while ensuring that the payload oscillations are kept minimum. An Integrated Decision Controller (IDC) is proposed, that integrates the optimal tracking control law given by a centralized Model Predictive Controller with safety-critical constraints provided by the Exponential Control Barrier Functions. The entire payload-UAV system is enclosed by a safe convex hull boundary, and the IDC ensures that no obstacle enters this boundary. To evaluate the performance of the IDC, the results for a numerical simulation as well as a high-fidelity \texttt{Gazebo} simulation are presented. An ablation study is conducted to analyze the robustness of the proposed IDC against practical dubieties like noisy state values, relative obstacle safety margin, and payload mass uncertainty. The results clearly show that the IDC achieves both trajectory tracking and obstacle avoidance successfully while restricting the payload oscillations within a safe limit.
\end{abstract}


\section{Introduction}

Recent advancements in Unmanned Aerial Vehicles (UAV) and sensor technology has created a widespread interest in their ability to solve payload transportation problems in the logistics sector. Collaborative interaction of UAVs can efficiently solve many transportation problems, especially when there are requirements for delivering versatile payloads that vary in size. This is because utilizing multiple UAVs to lift payloads are economically feasible and efficient, as opposed to utilizing a single UAV of a bigger form factor. Industrial missions\cite{9136895}, military expeditions\cite{9274974} and medical search-rescue operations \cite{arnold2018search} demand that the UAVs have to ensure proper delivery of the payload along a set pre-planned global trajectory, while at the same time, preventing collisions with the local dynamic obstacles present around it and maintaining minimal oscillations of the payload throughout the course. However, when there are multiple UAVs interacting with one another to transport a payload, existing path planning algorithms\cite{9517336} fall short, as they don't consider the kinodynamic-maneuverability of the system and it's highly coupled complex dynamics.

Majority of existing literature \cite{villa2020survey}, \cite{klausen2015cooperative}, \cite{liu2021analysis} models the payload suspension as a point mass cable suspension. This results in simplified system dynamics, which is generally unrealistic in situations where there are requirements to transport heavy payloads. In \cite{lee2015collision}, a payload suspension model is used, and formation control of UAVs are considered to negotiate a tight passage. However formation control is not practically possible for rigid link payload suspension with minimal oscillation constraints. Moreover, additional mechanisms like rack-pinion attachment points must be employed during formation control when rigid link suspensions are employed.  

In \cite{pizetta2019avoiding}, \cite{pizetta2018control} the payload is modelled as an external disturbance rather than a point mass. The direction of this external disturbance force is along the cable connecting the payload and the UAV. This may lead to suboptimal results, as the control law cannot leverage the benefits of knowing the actual dynamics of the payload. With this approach, it is also hard to ensure bounds on the states of the payload such as keeping the oscillations under certain threshold. Further, Artificial Potential Fields (APFs) are considered for the purpose of obstacle avoidance in \cite{pizetta2019avoiding}. However, APFs can have limitations in certain scenarios like passing between closely spaced obstacles, exhibiting oscillations or being trapped in local minima as discussed in \cite{koren1991potential}. Moreover, Control Barrier Functions (CBFs) are more suited for obstacle avoidance than APFs as discussed in \cite{singletary2020comparative}.
% Existing literature works on collaborative payload transport address only certain aspects of the aforementioned requirements. In \cite{pizetta2019avoiding}, the payload is treated as a point mass, connected by supporting cables. The dynamics is limited to a 2D plane passing through the two UAVs and the point mass payload. Further, Artificial Potential Fields (APFs) are used for the purpose of obstacle avoidance, which can have limitations in certain scenarios like passing between closely spaced obstacles, exhibiting oscillations or being trapped in local minima as discussed in \cite{koren1991potential}. Formation control of UAVs is discussed in \cite{lee2015collision} where the UAVs try to negotiate a tight passage, by changing their formations, while the point mass payload continues to track a set trajectory. A geometric controller is developed for this purpose. However, changing formation is only possible with a point mass suspension; for rigid payload suspension, the UAVs are generally attached at a fixed point on the payload via cables / rigid links and thus formation control becomes infeasible without causing payload oscillations. Furthermore, the maneuvers executed by the UAVs during formation control results in exceedingly high control input rates and values that can be practically infeasible. 

 In contrast, \cite{gimenez2018multi} models the cable-payload suspension as a series of mass-spring-damper system, thus taking into account the elastic nature of the cable. A null-space based controller is then designed for the purpose of trajectory tracking and obstacle avoidance. While cable suspension of light, point mass payload may be desirable in some circumstances, they aren't desirable when the payload is heavy, as they may cause undesirable uncontrollable oscillations due to the cable elasticity. Further, with cable suspension there can exist a point in the course of the mission where the cables can become slack, causing the payload to become massless, or even accelerate upwards. Controlling the states of the payload in these situations becomes complicated and may not be feasible in practical situations where the exact states of the payload like its relative position and linear acceleration are not known, but needs to be estimated based on the cable orientation and the tension in it. It can also lead to singularities in the dynamics of the system as discussed in \cite{sreenath2013dynamics}, resulting in extremely complex control laws.

Even though the aforementioned works address the problem of cooperative payload transport, it is evident that there is a need to model heavy payload as a rigid body, with rigid link suspension, that enables easy control over the states of the payload. Moreover, safety-critical payload transport must be ensured in a cluttered environment, with minimum deviation from the preset global trajectory while avoiding the obstacles. 

In this paper, the payload-UAV system is modelled end-to-end, keeping in mind the necessity to control the states of the payload throughout the course of the mission. The dynamics of the system is modelled using Lagrangian mechanics as discussed in \cite{lee2017geometric}, and rigid, massless links are used in place of cable suspension, as demonstrated in \cite{wehbeh2020distributed}. This makes it relatively easier to calculate the payload states at all times. Moreover, using rigid links provides complete control authority to the UAVs to handle the payload as required. To ensure accurate tracking of the payload, a centralized Model Predictive Controller (MPC) is employed. As of now, centralized MPC is shown to outperform distributed MPC at a lower sampling rate in \cite{wehbeh2020distributed}. Furthermore, the major advantage of using MPC based controllers is it's inherent ability to take into consideration the state and control input constraints. The concept of receding horizon also ensures that future uncertainties and requirements are taken into consideration while obtaining the control law at the current time step. Inspired by the application of control barrier functions to safety-critical systems, an obstacle avoidance controller is developed for the entire payload-UAV system using Exponential Control Barrier Functions (ECBFs). Barrier functions naturally provide safety-critical constraints that can be leveraged in an optimization framework. The \textit{Integrated Decision Controller} fuses the optimal tracking control law as provided by the MPC into an optimization-based obstacle avoidance decision controller. The IDC is designed in a way that allows any trajectory tracking controller to be used with the obstacle avoidance controller.

%To summarize, the main contributions of this paper is to develop an two-stage controller that guarantees obstacle avoidance while tracking a desired preset trajectory. The obstacle avoidance controller is based on the concept of ECBFs that take in nonlinear dynamical systems of higher relative degree, as discussed in Section \ref{subsec:ECBF}.

The rest of the paper is organized as follows: Section \ref{sec:sysModel} covers the system description and state space formulation. Section \ref{sec:obscon} covers the design and development of the Integrated Decision controller for the payload-UAV system, with brief discussion on Model Predictive Control and Exponential Barrier Functions. In Section \ref{sec:simres}, the proposed controller is evaluated, and the results for numerical simulations as well as a \texttt{Gazebo} simulation are presented along with an ablation study demonstrating the robustness of the IDC. Finally, conclusions and possibilities for future work is discussed in Section \ref{sec:conclusion}.

\begin{figure}
    \centering
    \begin{tikzpicture}[scale=1.0, every node/.style={transform shape}]
    %\draw[fill=gray] (0,0) ellipse (7pt and 3.5pt) node(prop11){};
    %\draw[fill=gray] (1.0, 0) ellipse (7pt and 3.5pt) node(prop12) {};
    
    %\draw[thin, -] (prop11) -- (prop12);
    
    %Quadrotor1, top left
    \node[ellipse, draw, fill = gray!60, minimum width = 0.55cm, minimum height = 0.08cm] (prop11) at (0,1) {};
    \node[ellipse, draw, fill = gray!60, minimum width = 0.55cm, minimum height = 0.08cm] (prop12) at (1,1) {};
    \node[ellipse, draw, fill = gray!60, minimum width = 0.55cm, minimum height = 0.08cm] (prop13) at (0.2,0.5) {};
    \node[ellipse, draw, fill = gray!60, minimum width = 0.55cm, minimum height = 0.08cm] (prop14) at (1.2,0.5) {};
    \draw[line width=0mm] (prop11) -- (prop14);
    \draw[line width=0mm] (prop12) -- (prop13);
    
    %Quadrotor2, top right
    \node[ellipse, draw, fill = gray!60, minimum width = 0.55cm, minimum height = 0.08cm] (prop21) at (3+0.5,0.7) {};
    \node[ellipse, draw, fill = gray!60, minimum width = 0.55cm, minimum height = 0.08cm] (prop22) at (3+1.5,0.7) {};
    \node[ellipse, draw, fill = gray!60, minimum width = 0.55cm, minimum height = 0.08cm] (prop23) at (3+0.2,0.2) {};
    \node[ellipse, draw, fill = gray!60, minimum width = 0.55cm, minimum height = 0.08cm] (prop24) at (3+1.2,0.2) {};
    \draw[line width=0mm] (prop21) -- (prop24);
    \draw[line width=0mm] (prop22) -- (prop23);
    
    \draw[-latex] (3.85, 0.45) -- (4.1, -0.5);
    \node at (5.2, -0.4) {$F_j = \mathcal{F}_j\pmb{R}_jb_{j3}$};
    
    %Quadrotor3, bottom right
    \node[ellipse, draw, fill = gray!60, minimum width = 0.55cm, minimum height = 0.08cm] (prop31) at (4+0.,-1.5) {};
    \node[ellipse, draw, fill = gray!60, minimum width = 0.55cm, minimum height = 0.08cm] (prop32) at (4+1.,-1.5) {};
    \node[ellipse, draw, fill = gray!60, minimum width = 0.55cm, minimum height = 0.08cm] (prop33) at (4+0,-1) {};
    \node[ellipse, draw, fill = gray!60, minimum width = 0.55cm, minimum height = 0.08cm] (prop34) at (4+1,-1) {};
    \draw[line width=0mm] (prop31) -- (prop34);
    \draw[line width=0mm] (prop32) -- (prop33);
    
    %Quadrotor4, bottom left
    \node[ellipse, draw, fill = gray!60, minimum width = 0.55cm, minimum height = 0.08cm] (prop41) at (-0.5+0.2,-1.9) {};
    \node[ellipse, draw, fill = gray!60, minimum width = 0.55cm, minimum height = 0.08cm] (prop42) at (-0.+0.5,-1.5) {};
    \node[ellipse, draw, fill = gray!60, minimum width = 0.55cm, minimum height = 0.08cm] (prop43) at (-0.7+0,-1.4) {};
    \node[ellipse, draw, fill = gray!60, minimum width = 0.55cm, minimum height = 0.08cm] (prop44) at (-0.5+0.6,-1.) {};
    \draw[line width=0mm] (prop41) -- (prop44);
    \draw[line width=0mm] (prop42) -- (prop43);
    
    %Payload
    \node[trapezium, draw, minimum width=3cm, trapezium left angle=120, trapezium right angle=60, line width=0mm] at (2.5,-3) {};
    %\node[trapezium, draw, minimum width=3cm, trapezium left angle=120, trapezium right angle=60, line width=0mm] at (2.5,-3.6) {};
    \draw[line width=0mm] (1.0, -2.3) -- (1.0, -2.9);
    %\draw[line width=0mm] (3.2, -2.3) -- (3.2, -2.9);
    \draw[line width=0mm] (1.5+0.3, -4.3+0.6) -- (1.5+0.3, -4.9+0.6);
    \draw[line width=0mm] (3.7+0.3, -4.3+0.6) -- (3.7+0.3, -4.9+0.6);
    
    \draw[line width=0mm] (1.0, -2.9) -- (1.8, -4.3) -- (4.0, -4.3);
    
    %Links 1-4
    \draw[line width=0mm] (1.0, -2.3) -- (0.6, 0.75);
    \draw[line width=0mm] (3.2, -2.3) -- (3.85, 0.45);
    \draw[line width=0mm] (1.8, -3.7) -- (-0.1, -1.45);
    \draw[line width=0mm] (4.0, -3.7) -- (4.5, -1.25);
    
    \draw [-latex, thick, shorten >= 2.00cm] (0.6, 0.75) -- (1.0, -2.3);
    \draw [-latex, thick, shorten >= 2.00cm] (3.85, 0.45) -- (3.2, -2.3);
    \draw [-latex, thick, shorten >= 2.00cm] (-0.1, -1.45) -- (1.8, -3.7);
    \draw [-latex, thick, shorten >= 1.50cm] (4.5, -1.25) -- (4.0, -3.7);
    
    \draw[line width=0mm, densely dashed] (2.5,-3.3) -- (4.0, -3.7);
    \draw[line width=0mm, densely dashed] (2.5,-3.3) -- (1.8, -3.7);
    \draw[line width=0mm, densely dashed] (2.5,-3.3) -- (1.0, -2.3);
    \draw[line width=0mm, densely dashed] (2.5,-3.3) -- (3.2, -2.3);
    
    \node at (2.5, -3.3) [circle,fill,inner sep=1.5pt]{};
    
    \node at (1.9, -2.6) {$p_i$};
    \node at (1.5, 0.0) {$q_i \in S^2$};
    \node at (1.1, -1.) {$l_i$};
    \node at (-0.6, 0.5) {$m_i, \pmb{J}_i$};
    \node at (0.9, -4.0) {$m_0, \pmb{J}_0$};
    \node at (0.0, 1.5) {$(r_i, \pmb{R}_i) \in SE(3)$};
    \node at (3.0, -4.7) {$(r_0, \pmb{R}_0) \in SE(3)$};
    
    \draw[<-] (-0.7, -5.2) -- (-1.2, -5.0);
    \draw[->] (-1.2, -5.0) -- (-1.2, -5.6);
    \draw[->] (-1.2, -5.0) -- (-0.8, -4.6);
    
    \node at (-0.6, -4.6) {\tiny x};
    \node at (-0.6, -5.2) {\tiny y};
    \node at (-1.3, -5.7) {\tiny z};
    
    \end{tikzpicture}
    
    \caption{A Schematic diagram of $N$ UAVs transporting a rigid body, attached via massless rigid links present in the NED coordinate frame is shown. The black dot at the center of the payload denotes its center of mass.}
    \label{fig:schDiag}
    
\end{figure}
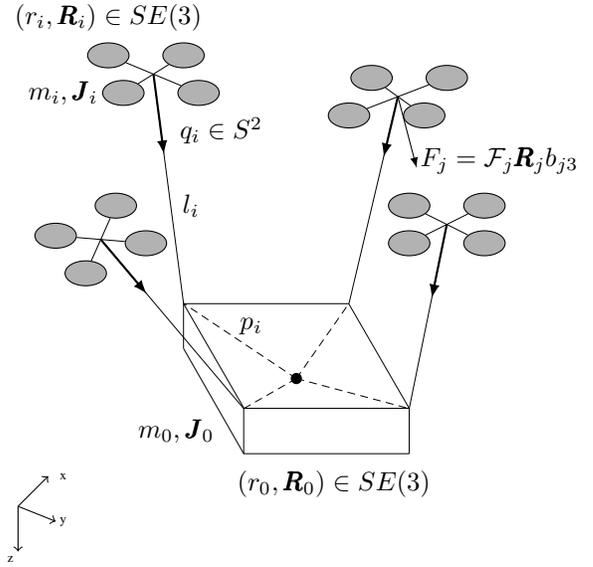

\section{System Model}
\label{sec:sysModel}
In this section, a brief description of the equations of motion for the payload-UAV system is provided. The derivation of the system dynamics parallels the discussion presented in \cite{lee2017geometric} except that the equations of motion are resolved in the payload frame of reference \cite{wehbeh2020distributed}.
\subsection{Payload-UAV System Description}
There are $N$ UAVs that are collaboratively transporting a rigid payload, as shown in Fig. \ref{fig:schDiag}. The UAVs are attached to the payload via massless rigid links. The rigid links are connected to the payload and the UAVs via spherical joints. This way, the attitude dynamics of the UAVs are decoupled from the dynamics of the rest of the system.

Throughout this paper, the NED (North, East, Down) coordinate system is used, with the positive $z$-axis pointing downward. The direction for positive $x$-, $y$- axis are then chosen on the basis of the right hand rule. The basis vectors for the inertial frame and the body-fixed frames are denoted by $\left\{  e_1, e_2, e_3\right\}$ and $\left\{  b_{i1}, b_{i2}, b_{i3}\right\}$ respectively. The variables that refer to the payload are subscripted by $0$, and the variables that refer to the $i^{th}$-UAV is subscripted by $i$.

The location of the center of mass of the payload in the inertial frame is denoted by $r_0 \in \mathbb{R}^3$, and the location of the center of mass of the $i^{th}$-UAV in the inertial frame is denoted by $r_i \in \mathbb{R}^3$. The position vector of the attachment point of the rigid link of length $l_i$ corresponding to the $i^{th}$-UAV in the payload frame is denoted by $p_i \in \mathbb{R}^3$. The unit vector along the rigid link corresponding to the $i^{th}$-UAV, pointing towards the payload is denoted by $q_i \in \mathbb{S}^2$, where $\mathbb{S}^2 = \left\{ q_i \in \mathbb{R}^3 \ | \ \Vert q_i \Vert_{_2} = 1 \right\}$. The attitude of the payload is characterized by $\pmb{R}_0 \in SO(3)$, the rotation matrix that rotates a vector in the payload-fixed frame to the inertial frame. Similarly, the attitude of the $i^{th}$-UAV is characterized by $\pmb{R}_i \in SO(3)$, the rotation matrix that rotates a vector in the $i^{th}$-UAV fixed frame to the inertial frame. Let $\pmb{J}_0 \in \mathbb{R}^{3 \times 3}$ denote the inertia matrix of the payload with mass $m_0$, and $\pmb{J}_i \in \mathbb{R}^{3 \times 3}$ denote the inertia matrix of the $i^{th}$-UAV with mass $m_i$. 

%The angular velocities of the rigid massless links corresponding to the $i^{th}$-UAV is denoted by $\Omega_i \in \mathbb{R}^3$, satisfying the orthogonal condition $q_i \cdot \Omega_i = 0$. The angular velocities of the payload and the $i^{th}$-UAV is denoted by $\omega_0, \omega_i \in \mathbb{R}^3$ respectively.

The force vector $F_i \in \mathbb{R}^3$ generated by the $i^{th}$-UAV in the inertial frame is denoted by $F_i = \mathcal{F}_{i}\pmb{R}_i b_{i3}$, where $\mathcal{F}_{i} \in \mathbb{R}$ denotes the thrust value. In addition to this, the torque vector generated by the $i^{th}$-UAV in the body-fixed frame is denoted by $\tau_i \in \mathbb{R}^3$. Thus, the control inputs to the payload-UAV system is given by $\left\{ \mathcal{F}_i, \tau_i \right\}, i \in \{ 1, .. , N \}$. 

\subsection{Payload-UAV System Dynamics}
Based on the notations and the system description presented above, the equations of motion are derived in the following section. As the connecting links are rigid and massless, the position vector $r_i$ of the $i^{th}$-UAV can be directly inferred as:
\begin{align}
    r_i = r_0 + \pmb{R}_0\left( p_i - l_iq_i \right), \ \ \ i = 1,\ldots,N
\end{align}
The kinematic equations for the payload-UAV system are:
\begin{align}
    \Dot{r}_0 &= \pmb{R}_0 v_0 \label{eq:kinBeg}\\
    \pmb{\Dot{R}}_0 &= \pmb{R}_0 \omega_0^\times \\
    \Dot{r}_i &= \Dot{r}_0 + \pmb{R}_0 \left( \omega_0^\times p_i - l_i\Omega_i^\times q_i \right) \\
    \Dot{q}_i &= \left( \Omega_i - \omega_0 \right)^\times q_i \\
    \pmb{\Dot{R}}_i &= \pmb{R}_i \omega_i^\times,\ \ \ \ \ \ \ \ \ \ \ \ \ \ \ \ \ \ i=1,\ldots,N \label{eq:kinEnd}
\end{align}

where the hat map $(\cdot)^\times : \mathbb{R}^3 \rightarrow SO(3)$ denotes the skew-symmetric operator, $v_0 \in \mathbb{R}^3$ denotes the velocity of the payload in the payload-fixed frame, $w_0, w_i \in \mathbb{R}^3$ denotes the angular velocities of the payload and the $i^{th}$ UAV respectively and $\Omega_i \in \mathbb{R}^3$ denotes the angular velocities of the $i^{th}$ link. Note that the orthogonal condition $q_i \cdot \Omega_i = 0$ must be satisfied for all time.

The kinetic energy $\mathcal{T}$ and the potential energy $\mathcal{U}$ of the system are given by:
\begin{align}
    \begin{split}
        \mathcal{T} = {}& \frac{1}{2} m_0 \Vert \Dot{r}_0 \Vert^2_{_2} + \frac{1}{2} \omega_0 \cdot \pmb{J}_0\omega_0 \\ 
        & + \sum_{i=1}^N \left( \frac{1}{2} m_i \Vert \Dot{r}_i \Vert^2_{_2} + \frac{1}{2} \omega_i \cdot \pmb{J}_i \omega_i \right)
    \end{split} \\
    \begin{split}
        \mathcal{U} = {}& -m_0 g \mathbb{k} \cdot r_0 - \sum_{i=1}^N m_ig\mathbb{k} \cdot r_i
    \end{split}
\end{align}
where $\mathbb{k} = \left[ 0 \ 0 \ 1 \right]^T$. To derive the equations of motion of the payload-UAV system, the \textit{Lagrangian} $\mathcal{L}$ of the system must be obtained:
\begin{align}
         \mathcal{L} = {}& \mathcal{T} - \mathcal{U}
\end{align}
 Once the Lagrangian $\mathcal{L}$ is formulated, the Euler-Lagrange equations can be obtained from the \textit{Lagrange-d’Alembert Principle} as described in the Appendix section of \cite{lee2017geometric} and are given as follows:
\begin{align}
    \frac{d}{dt} \pmb{D}_{\Dot{r}_0} \mathcal{L} &- \pmb{D}_{r_0} \mathcal{L} = \sum_{i=1}^N F_i \\
    \frac{d}{dt} \pmb{D}_{\omega_0} \mathcal{L} + \omega_0^\times \pmb{D}_{\omega_0}\mathcal{L} &- \pmb{d}_{\pmb{R}_0} \mathcal{L} = \sum_{i=1}^N p_i^\times \pmb{R}_0^T F_i \\
    q_i^\times\frac{d}{dt}\pmb{D}_{\Dot{q}_i}\mathcal{L} &- q_i^\times \pmb{D}_{q_i}\mathcal{L} = -l_i q_i^\times F_i \\
    \frac{d}{dt} \pmb{D}_{\omega_i} \mathcal{L} &+ \omega_i^\times \pmb{D}_{\omega_i} \mathcal{L} = \tau_i
\end{align}
where $\pmb{D}_{a} \mathcal{L}$ is the derivative of the Lagrangian $\mathcal{L}$ w.r.t the vector $a$, and $\pmb{d}_{\pmb{R}_0}\mathcal{L}$ are the \textit{left-trivialized derivatives} of the Lagrangian $\mathcal{L}$ \cite{lee2017geometric}. Substituting for the derivatives of the Lagrangian, and simplifying the equations leads to the equations of motion of the payload-UAV system \cite{wehbeh2020distributed}:
\begin{gather}
    \begin{split} \shoveleft
         m_T \left( \Dot{v}_0 + \omega_0^\times v_0\right) + \sum_{i=1}^N m_i \left( -p_i^\times\Dot{\omega}_0 + l_iq_i^\times\Dot{\Omega}_i + l_iq_i^\times\omega_0^\times\Omega_i \right) \\
        +\sum_{i=1}^N\left( m_i \left( \omega_0^\times \right)^2 p_i + m_il_i\Vert \Omega_i \Vert_{_2}^2 q_i \right) = \pmb{R}_0^T\left(m_Tg\mathbb{k} + \sum_{i=1}^N F_i \right) \label{eq:EqBeg}
    \end{split} \\
    \begin{split}
         \sum_{i=1}^N m_ip_i^\times\left( \Dot{v}_0 + \omega_0^\times v_0 + l_iq_i\Dot{\Omega}_i  + l_i\Vert \Omega_i \Vert_{_2}^2q_i + l_iq_i^\times\omega_0^\times\Omega_i \right) + \\ \pmb{\Bar{J}}_0\Dot{\omega}_0 + \omega_0^\times \pmb{\Bar{J}}_0\omega_0 = \sum_{i=1}^N p_i^\times\pmb{R}_0^T\left( F_i + m_ig\mathbb{k} \right) \ \ \ \ \ \ \ \ \ \hfill
    \end{split} \\
    \begin{split}
        m_il_i^2\left( \Dot{\Omega}_i + \omega_0^\times\Omega \right) - m_il_iq_i^\times\left( \Dot{v}_0 + \omega_0^\times v_0 \right) + m_il_iq_i^\times p_i^\times\Dot{\omega}_0 \\
        -m_il_iq_i^\times\left( \omega_0^\times \right)^2 p_i = -l_iq_i^\times\pmb{R}_0^T\left( F_i + m_ig\mathbb{k} \right) \ \ \ \ \ \ \ \hfill
    \end{split} \\
    \begin{split}
         \pmb{J}_i\Dot{\omega}_i + \omega_i^\times\pmb{J}_i\omega_i = \tau_i
        \label{eq:EqEnd}
    \end{split}
\end{gather}

where $m_T = \left(m_0 + \sum_{i=1}^Nm_i\right)$ is the total mass of the UAVs with the payload and $\pmb{\Bar{J}}_0 = \left(\pmb{J}_0 - \sum_{i=1}^Nm_i\left( p_i^\times \right)^2\right)$.

\subsection{State-Space Representation and Model Linearization}
The equations of motion described earlier can be rearranged, and a state-space representation of the payload-UAV system can be obtained. Consider the state vector $x \in \mathbb{R}^{12 + 16N}$ as follows:
\begin{align}
    x = \left[\underbrace{r_0^T \ v_0^T \ \Theta_0^T \ \omega_0^T}_\text{payload $\in \mathbb{R}^{12}$} \ \underbrace{q_i^T \ \Omega_i^T \ \Theta_i^T \ \omega_i^T}_\text{$i^{th}$ UAV $\in \mathbb{R}^{16N}$} \right]^T, \ i \in \{1,..., N \}
    \label{eq:stateqn}
\end{align}
where $\Theta_0$, $\Theta_i \in \mathbb{R}^3$ are the ZYX Euler angle parametrization of $\pmb{R}_0$ and $\pmb{R}_i$ respectively. The state-space equations for $\Dot{x}_a = \left[\Dot{r}_0^T \  \Dot{\Theta}_0^T \ \Dot{q}_i^T \ \Dot{\Theta}_i^T\right]^T$ can be easily obtained from Eq. (\ref{eq:kinBeg}) - (\ref{eq:kinEnd}). The equations for $\Dot{x}_b = \left[ \Dot{v}_0^T \ \Dot{\omega}_0^T \ \Dot{\Omega}_i^T \ \Dot{\omega}_i^T \right]^T$ can be obtained by rearranging equations Eq. (\ref{eq:EqBeg}) - (\ref{eq:EqEnd}) and is given by:
\begin{align}
    \Dot{x}_b = \pmb{P}^{-1}\pmb{Q} 
\end{align}
where $\pmb{P} =$
\begin{equation}
\resizebox{.94\hsize}{!}{$
\left[\begin{array}{cc|ccc}
m_{T} \pmb{I}_{3} & -\sum_{i=1}^{N} m_{i} p_{i}^{\times} & m_{1} l_{1} q_{1}^{\times} & \ldots & m_{N} l_{N} q_{N}^{\times} \\
\sum_{i=1}^{N} m_{i} p_{i}^{\times} & \pmb{\Bar{J}}_{0} & m_{1} l_{1} p_{1}^{\times} q_{1}^{\times} & \ldots & m_{N} l_{N} p_{N}^{\times} q_{N}^{\times} \\
\hline m_{1} l_{1} q_{1}^{\times} & m_{1} l_{1} q_{1}^{\times} p_{1}^{\times} & m_{1} l_{1}^{2} \pmb{I}_{3} & \ldots & \pmb{0}_3 \\
\vdots & \vdots & \vdots & \vdots & \vdots \\
-m_{N} l_{N} q_{N}^{\times} & m_{N} l_{N} q_{N}^{\times} p_{N}^{\times} & \pmb{0}_3 & \ldots & m_{N} l_{N}^{2} \pmb{I}_{3}
\end{array}\right]$}
\end{equation}

and $\pmb{Q} =$

\begin{equation}
\left[\begin{array}{c}
-m_{T} \omega_{0}^{\times} v_{0}-\sum_{i=1}^{N}\left\{m_{i}\left(\omega_{0}^{\times}\right)^{2} p_{i}+m_{i} l_{i}\left\Vert\Omega_{i}\right\Vert_{_2}^{2} q_{i}\right. \\
\left.+m_{i} l_{i} q_{i}^{\times} \omega_{0}^{\times} \Omega_{i}\right\}+m_{T} g \pmb{R}_{0}^{T} \mathbb{k}+\sum_{i=1}^{N} \pmb{R}_{0}^{T} F_{i} \\
\hline-\omega_{0}^{\times} \Bar{\pmb{J}}_{0} \omega_{0}-\sum_{i=1}^{N} m_{i}\left\{p_{i}^{\times} \omega_{0}^{\times} v_{0}+l_{i} p_{i}^{\times} q_{i}^{\times} \omega_{0}^{\times} \Omega_{i}\right. \\
\left.+l_{i} p_{i}^{\times}\left\Vert\Omega_{i}\right\Vert_{_2}^{2} q_{i}\right\}+\sum_{i=1}^{N} p_{i}^{\times} \pmb{R}_{0}^{T}\left(F_{i}+m_{i} g \mathbb{k}\right) \\
\hline m_{1} l_{1}\left\{q_{1}^{\times}\left(\omega_{0}^{\times}\right)^{2} p_{1}-l_{1} \omega_{0}^{\times} \Omega_{1}+q_{1}^{\times} \omega_{0}^{\times} v_{0}\right\} \\
-l_{1}q_1^{\times} \pmb{R}_{0}^{T}\left(F_{1}+m_{1} g \mathbb{k}\right) \\
\hline
\vdots \\
\hline \\
m_{N} l_{N}\left\{q_{N}^{\times}\left(\omega_{0}^{\times}\right)^{2} p_{N}-l_{N} \omega_{0}^{\times} \Omega_{N}+q_{N}^{\times} \omega_{0}^{\times} v_{0}\right\} \\
-l_{N} q_{N}^{\times} \pmb{R}_{0}^{T}\left(F_{N}+m_{N} g \mathbb{k}\right)
\end{array}\right]
\end{equation}

where $\pmb{0}_n \in \mathbb{R}^{n\times n}$ is the $n\times n$ zero matrix and $\pmb{I}_n \in \mathbb{R}^{n\times n}$ is the $n\times n$ identity matrix. The control input to the system is defined by the vector $u = \left[ \mathcal{F}_1 \ \tau_1^T \ .. \ \mathcal{F}_N \ \tau_N^T \right]^T \in \mathbb{R}^{4N}$. The equilibrium points of the system $\left\{x_e, u_e \right\}$ satisfy the condition $f\left(x_e, u_e \right) = 0$. One such equilibrium point ($x_e, u_e$) for the payload-UAV system is given in Eq. \ref{eq:EqPt}, and occurs when all the links are vertical, and the orientation of the payload and all the UAVs are zero, thus producing zero torques. In addition to this, the thrust generated by the UAVs must balance the overall weight they are subjected to, maintaining net zero acceleration:
\begin{subequations}
\begin{align}
    x_e = \left[ r_0^T \ \mathbb{0}^T \ \mathbb{0}^T \ \mathbb{0}^T \ \mathbb{k}_i^T \ \mathbb{0}_i^T \ \mathbb{0}_i^T \ \mathbb{0}_i^T \right]^T , i \in \{1, .., N \} \\
    u_e = \left[ \mathcal{F}_{e_i} \ \mathbb{0}_i^T \right]^T, i \in \{1, .., N \} \ \ \ \ \ \ \ \ \ \ \ \ \
\end{align}
\label{eq:EqPt}
\end{subequations}
where $\mathbb{0} \in \mathbb{R}^3$ is the zero vector and $\mathcal{F}_{e_i} = \left(m_i + m_0/4\right)g$. The non-linear state-space equations can then be linearized about the equilibrium point $\left( x_e, u_e \right)$. The resultant discretized equations of motion at time step $k$ is given by:

\begin{subequations}
\begin{align}
    \Delta x_{k+1} &= \pmb{A}\Delta x_k + \pmb{B}\Delta u_k \\
    \Delta y_k &= \pmb{C}\Delta x_k + \pmb{D}\Delta u_k
\end{align}
\label{eq:disSys}
\end{subequations}
where $\pmb{A} \in \mathbb{R}^{n\times n}$ is the state matrix\footnoteref{ft:AB}, $\pmb{B} \in \mathbb{R}^{n\times m}$ is the input matrix\footnote{\label{ft:AB} Due to their high dimensionality, the matrices are uploaded as .csv files, and can be viewed \href{https://drive.google.com/drive/folders/1ZIwZ4higpC-BZ5isK-Gr1uozPNqTscE7?usp=sharing}{here}.}, $\pmb{C} \in \mathbb{R}^{p\times n}$ is the output matrix and $\pmb{D} \in \mathbb{R}^{p\times m}$ is the input feedforward matrix. The discrete state vector $\Delta x_k \in \mathbb{R}^n$ and the control input vector $\Delta u_k \in \mathbb{R}^m$ is obtained as $\Delta x_k = x_k - x_e$, $\Delta u_k = u_k - u_e$ respectively. Due to the principle of receding horizon, it is assumed implicitly that the input $u_k$ cannot affect the output $y_k$ at the same time. Thus, throughout the discussion that follows, it is assumed that $\pmb{D} = \pmb{0}$.

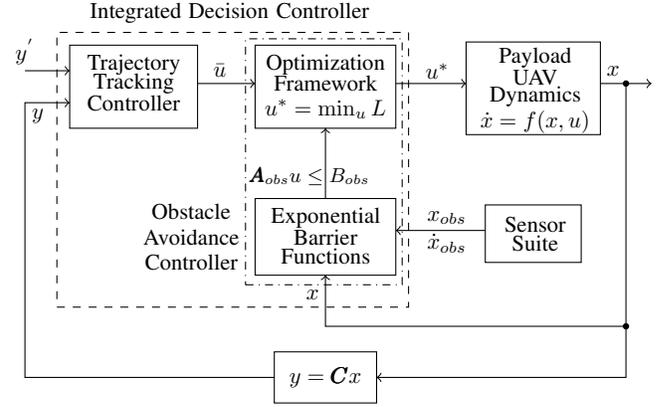
\begin{figure}
    
    \begin{tikzpicture}[scale=0.85, every node/.style={transform shape}]
        \draw (0, 0.1) rectangle (2, 1.4);
        \node at (1, 1.1) {Trajectory};
        \node at (1, 0.8) {Tracking};
        \node at (1, 0.5) {Controller};
        
        \draw[->] (2, 0.8) -- (2.9, 0.8); 
        \node at (2.35, 1.0) {$\Bar{u}$};
        
        \draw (2.9, 0.1) rectangle (5.1, 1.4);
        \node at (4, 1.1) {Optimization};
        \node at (4, 0.8) {Framework};
        \node at (4, 0.4) {$u^* = \min_u L$};
        
        \draw[dashed] (-0.2, -2.7) rectangle (5.3, 1.6);
        \node at (2.5, 1.9) {Integrated Decision Controller};
        
        \draw (2.9, -2.2) rectangle (5.1, -1.0);
        \node at (4, -1.3) {Exponential};
        \node at (4, -1.6) {Barrier};
        \node at (4, -1.9) {Functions};
        
        \draw [->] (4, -1.0) -- (4, 0.1);
        \node at (3.75, -0.65) {\small $\pmb{A}_{obs}u \leq B_{obs}$};
        
        \draw[dashdotted] (2.75, -2.35) rectangle (5.2, 1.5);
        \node at (1.9, -1.2) {Obstacle};
        \node at (1.9, -1.6) {Avoidance};
        \node at (1.9, -2.0) {Controller};
        
        \draw[->] (5.1, 0.8) -- (6.2, 0.8); 
        \node at (5.75, 1.0) {$u^*$};
        
        \draw (6.2, 0.0) rectangle (8.3, 1.45);
        \node at (7.28, 1.2) {Payload};
        \node at (7.28, 0.9) {UAV};
        \node at (7.28, 0.6) {Dynamics};
        \node at (7.28, 0.2) {$\dot{x} = f(x, u)$};
        
        \draw (6.5, -2.0) rectangle (8.0, -1.1);
        \node at (7.28, -1.4) {Sensor};
        \node at (7.28, -1.7) {Suite};
        
        \draw[->] (6.5, -1.5) -- (5.1, -1.5);
        \node at (5.9, -1.3) {$x_{obs}$};
        \node at (5.9, -1.7) {$\dot{x}_{obs}$};
        
        \draw[->] (8.3, 0.8) -- (9.1, 0.8);
        \node at (8.5, 1.0) {$x$};
        \node at (8.7, 0.8)[circle,fill,inner sep=1.pt] {};
        \draw[->] (8.7, 0.8) -- (8.7, -3.8) -- (4.8, -3.8);
        \draw (3.2, -4.2) rectangle (4.8, -3.4);
        \node at (4, -3.8) {$y = \pmb{C}x$};
        \draw[->] (3.2, -3.8) -- (-0.7, -3.8) -- (-0.7, 0.5) -- (0.0, 0.5);
        \draw[->] (-0.7, 1.0) -- (0.0, 1.0); 
        \node at (-0.7, 1.3) {$y^{'}$};
        \node at (-0.5, 0.3) {$y$};
        
        \node at (8.7, -3.0) [circle, fill, inner sep=1.pt] {};
        \draw[->] (8.7, -3.0) -- (4, -3.0) -- (4, -2.2);
        \node at (3.8, -2.5) {$x$};
        
    \end{tikzpicture}
    \caption{Schematic Diagram of the Integrated Decision Controller. The IDC (dashed box) comprises of a trajectory tracking controller and an obstacle avoidance controller (dash-dot box)}
    \label{fig:schCont}
\end{figure}

\section{Integrated Decision Controller design for safety-critical navigation}
\label{sec:obscon}

\begin{figure*}
\hspace{0.5cm}
\includegraphics[scale=0.35]{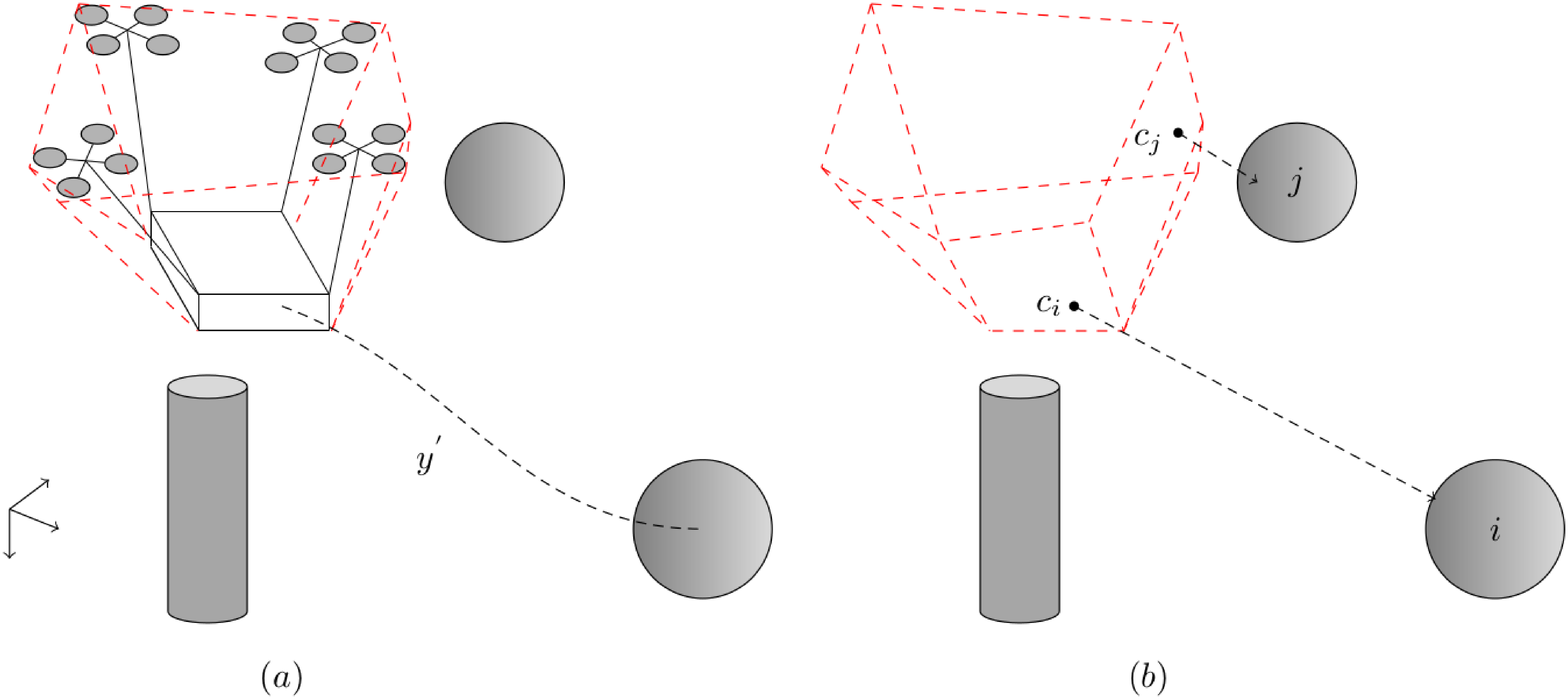}
    \caption{The payload must track the desired trajectory $y'$ amidst a cluttered environment. The cluttered environment consists of static and dynamic obstacles which are shown by spheres and cylinders. The formation of the convex hull (dashed lines) around the payload-UAV system is shown in (a). The convex hull encloses all the UAVs and the payload completely. In (b), the point $c_i, c_j$ are the points on the convex hull closest to the $i^{th}$ and the $j^{th}$ obstacle respectively.}
    
    \label{fig:cvxHull}
\end{figure*}

In this section, an integrated decision controller (IDC) is designed, that meets the requirements of trajectory tracking and obstacle avoidance. Particularly, the IDC must ensure that the payload oscillations are minimum, while tracking a set trajectory. Moreover, the IDC must be capable of providing safety critical decision control in order to avoid dynamic obstacles in the cluttered environment, while maintaining minimum deviation from the set trajectory. The schematic of the IDC is shown in Fig. (\ref{fig:schCont}).

The structure of IDC ensures powerful versatility in the sense that any tracking controller can be used in conjunction with the obstacle avoidance controller. In this work, a Model Predictive Controller is designed for the linear discretized system (\ref{eq:disSys}), with particular emphasis on the derivation of state and control input constraints. This way, the bounds on payload oscillations and tracking requirements can be automatically ensured in a single framework.

\subsection{Safety critical payload transport in a cluttered environment}

This work focuses on ensuring safety critical transportation of a payload in a cluttered environment. In particular, there must be no collision of the payload with any surrounding obstacles, and the payload oscillations must be restricted under a given limit:
\begin{subequations}
\begin{align}
    \left\vert \phi_0 \right\vert &\leq \delta_{\phi}, \\
    \left\vert \theta_0 \right\vert &\leq \delta_{\theta}, \\ 
    \left\vert \psi_0 \right\vert &\leq \delta_{\psi}\ \ \ \forall t > t_0
\end{align}
\label{eq:consOsc}
\end{subequations}
where $\Theta_0 = \left[\phi_0 \ \theta_0 \ \psi_0 \right]^T$ is the roll-pitch-yaw (rpy) of the payload, and $\delta_\phi, \delta_\theta, \delta_\psi > 0$ are small positive constants. The cluttered environment consists of static and dynamic obstacles, whose position $x_{obs}$ and velocity $\dot{x}_{obs}$ can be estimated by the onboard UAV sensor suite.

\subsection{Trajectory Tracking Controller}
A Linear Model Predictive Controller (LMPC) can be designed for the system (\ref{eq:disSys}), such that the payload tracks a given reference trajectory $y'$ as shown in in Fig. \ref{fig:cvxHull}(a). The LMPC minimizes the cost function $J$ at every time step $k$:
% \begin{subequations}
% \begin{align}
%     \Delta \Bar{U}_k = \argmin_{\Delta U_k} J(\Delta x_k, \Delta U_k) \ \ \ \ \ \ \ \ \ \\
%     \text{s.t} \ \ \left[\begin{array}{l} \pmb{M}_U \\ \pmb{M}_x\pmb{H} \end{array}\right]\Delta U_k \leq \left[\begin{array}{l} \ \ \ \ \ \Delta U_b \\ \Delta Z_b - \pmb{M}_x\pmb{F}\Delta x_k \end{array}\right] 
% \end{align}
% \label{eq:JMPC}
% \end{subequations}
\begin{multline}
    J\left( \Delta x_k, \Delta U_k  \right) = \left\Vert \pmb{\bar{C}}X_e + \Delta Y_k - Y'_k \right\Vert^2_{\pmb{\bar{Q}}} + \left\Vert \Delta U_k \right\Vert^2_{\pmb{\bar{R}}} \\
    + \left\Vert y_{k + N_p} - y'_{k + N_p} \right\Vert^2_{\pmb{Q}_f}\ \ \ \ \ \ \ \ \ \ \ \ \ \ \ \label{eq:JMPC}
\end{multline}
where $\bar{\pmb{C}} = \text{diag}(\pmb{C},..,\pmb{C})$, $\bar{\pmb{Q}} = \text{diag}(\pmb{Q},..,\pmb{Q})$, $\bar{\pmb{R}} = \text{diag}(\pmb{R},..,\pmb{R})$, $X_e = \left[x^T_e \ .. \ x^T_e\right]^T$, $\Delta Y_k = \left[ \Delta y^T_k \ .. \ \Delta y^T_{k+N_p - 1} \right]^T$ and $\Delta U_k = \left[ \Delta u^T_k,..,\Delta u^T_{k + N_p - 1}  \right]^T$ are defined over the prediction horizon $N_p$. The matrices $\pmb{Q}, \pmb{Q}_f \geq 0$ and $\pmb{R} > 0$ are the cost weighting matrices and $y_k = \pmb{C}\left(x_e + \Delta x_k\right)$. The matrix $\pmb{Q}_f$ is chosen to be equal to the solution of the discrete algebraic Riccati equation for the discrete system in Eq. (\ref{eq:disSys})\cite{ru2017nonlinear}. 

The cost function in Eq. \ref{eq:JMPC} is subject to linear state and control input constraints. The nonlinear constraints derived in \cite{wehbeh2020distributed} needs to be linearized about every time step along with the model. This can be computationally demanding in a real time implementation. A much simplified set of constraints can be derived for the discrete linear system in Eq. (\ref{eq:disSys}). Let the upper and lower bounds for the input be $u_{ub}$ and $u_{lb}$ respectively. Thus, $u_{lb} \leq u_e + \Delta u_{k+i} \leq u_{ub}$, which can also be expressed as 
\begin{align}
    \left[ \begin{array}{l} \pmb{I}_m \\ -\pmb{I}_m \end{array} \right]\Delta u_{k+i} \leq \left[ \begin{array}{l} \left( u_{ub} - u_e \right) \\ \left( u_e - u_{lb} \right) \end{array} \right]
\end{align}

Similarly, let the upper and lower bounds for the state be $x_{ub}$ and $x_{lb}$ respectively, for the set of states described by $\pmb{c}_zx_{k+i}$, where the matrix $\pmb{c}_z$ is a diagonal matrix, with ones for the states that require constraints, and zeros for the states that are free to take any values. The constraint $x_{lb} \leq \pmb{c}_z\left(x_e + \Delta x_{k+i}\right) \leq x_{ub}$ can be expressed as
\begin{align}
    \left[ \begin{array}{l} \pmb{c}_z \\ -\pmb{c}_z \end{array} \right]\Delta x_{k+i} \leq \left[ \begin{array}{l} \left( x_{ub} - \pmb{c}_z x_e \right) \\ \left( \pmb{c}_zx_e - x_{lb} \right) \end{array} \right]
\end{align}
The state vector $\Delta x_{k+i}$ can be recursively substituted in Eq. (\ref{eq:disSys}), to obtain $\Delta X_k = \left[ \Delta x^T_k \ ,.., \ \Delta x^T_{k + N_p - 1} \right]^T$ and thus, over the prediction horizon these constraint equations can be compactly written as:
\begin{align}
    \label{eq:inpCon}
    \pmb{M}_U \Delta U_k & \leq \Delta U_b \\
    \label{eq:stateCon}
    \pmb{M}_x\left( \pmb{F}\Delta x_k + \pmb{H}\Delta U_k \right) & \leq \Delta Z_b
\end{align}
where,
\begin{align*}
\begin{aligned}
    \pmb{H} &= \left[ \begin{array}{ccccc} \pmb{0} & & & & \\ \pmb{B} & \pmb{0} & & & \\ \pmb{AB} & \pmb{B} & \pmb{0} & & \\ \vdots & \vdots & \ddots & \ddots & \\ \pmb{A}^{N_p - 2}\pmb{B} & \pmb{A}^{N_p - 3}\pmb{B} & \ldots & \pmb{B} & \pmb{0}  \end{array} \right] \\
    \pmb{M}_U &= \left[ \begin{array}{ccc} \left[\begin{array}{l} \pmb{I}_m \\ -\pmb{I}_m \end{array}\right] & & \\ & \ddots & \\ & & \left[\begin{array}{l} \pmb{I}_m \\ -\pmb{I}_m \end{array}\right] \end{array} \right] \\
    \pmb{M}_x &= \left[ \begin{array}{ccc} \left[\begin{array}{l} \pmb{c}_z \\ -\pmb{c}_z \end{array}\right] & & \\ & \ddots & \\ & & \left[\begin{array}{l} \pmb{c}_z \\ -\pmb{c}_z \end{array}\right] \end{array} \right] \\
    \Delta U_b &= \left[ \begin{array}{l} \left[ \begin{array}{l} \left( u_{ub} - u_e \right) \\ \left( u_e - u_{lb}\right) \end{array} \right] \\ \ \ \ \ \ \ \ \ \ \vdots \\ \left[ \begin{array}{l} \left( u_{ub} - u_e \right) \\ \left( u_e - u_{lb} \right) \end{array} \right] \end{array} \right], \ \pmb{F} = \left[ \begin{array}{l} \pmb{I} \\ \pmb{A} \\ \pmb{A}^2 \\ \vdots \\ \pmb{A}^{N_p - 1} \end{array} \right]
\end{aligned}
\end{align*}
\begin{align*}
    \Delta Z_b &= \left[ \begin{array}{l} \left[ \begin{array}{l} \left( x_{ub} - \pmb{c}_zx_e \right) \\ \left( \pmb{c}_zx_e - x_{lb}\right) \end{array} \right] \\ \ \ \ \ \ \ \ \ \ \vdots \\ \left[ \begin{array}{l} \left( x_{ub} - \pmb{c}_zx_e \right) \\ \left( \pmb{c}_zx_e - x_{lb} \right) \end{array} \right] \end{array} \right], \ \ \ \ \ 
\end{align*}
Eq. (\ref{eq:stateCon}) can be rearranged, to obtain the constrain in terms of $\Delta U_k$, as in Eq. (\ref{eq:inpCon}). Once these constraints are obtained, the cost function in Eq. (\ref{eq:JMPC}) can be solved as:
\begin{subequations}
\begin{empheq}[box=\fbox]{align}
    \Delta \Bar{U}_k = \argmin_{\Delta U_k} J(\Delta x_k, \Delta U_k) \ \ \ \ \ \ \ \ \ \\
    \text{s.t} \ \ \left[\begin{array}{l} \pmb{M}_U \\ \pmb{M}_x\pmb{H} \end{array}\right]\Delta U_k \leq \left[\begin{array}{l} \ \ \ \ \ \Delta U_b \\ \Delta Z_b - \pmb{M}_x\pmb{F}\Delta x_k \end{array}\right]
\end{empheq}
\label{eq:J_MPC}
\end{subequations}
and the optimal tracking control input over $N_p$ horizon $\Bar{U}_k$ is given by $\bar{U}_k= U_e + \Delta \Bar{U}_k$, where $U_e = \left[ u_e^T \ldots u_e^T \right]^T$.
 
The constraints include bounds on the control input $u = \left[ \mathcal{F}_1 \ \tau_1^T \ldots \mathcal{F}_N \ \tau_N^T\right]^T$, and bounds on states such as the $z-$ coordinates of the payload and the UAVs, which cannot be positive (due to the NED convention) and the set of constraints in Eq. (\ref{eq:consOsc}). The reference trajectory $y^{'}$ contains the desired position of the payload, the desired orientation of the payload to be zero ensuring minimal oscillations and the desired orientation of links to be vertical, ensuring that the UAVs don't collide with each other, and the system stays close to the equilibrium point $\left( x_e, u_e\right)$ as described in Eq. (\ref{eq:EqPt}). Thus, the optimal tracking control input $\Bar{u}_k$ at the current step $k$ is obtained as $\bar{u}_k = [\pmb{I} \ \pmb{0} \ \ldots \ \pmb{0} ]\Bar{U}_k$.

\subsection{Obstacle Avoidance Controller}

The obstacles are dynamic in nature, and it is assumed that the position and velocity of the obstacle are available at every time step $k$. Due to the spatial structure of the payload-UAV system, it must be ensured that the rigid links and the propellers of the UAVs do not collide with the surrounding obstacles. This is ensured by constructing a \textit{safe} convex hull around the payload-UAV system such that it fully encloses it, as shown in Fig. \ref{fig:cvxHull}(a).

The collision avoidance problem is solved using the Exponential Control Barrier Functions (ECBFs) \cite{nguyen2016exponential}. In general, the control barrier function is defined over the state-space, and provides safety-critical constraints, that can be leveraged in an optimization framework. The ECBF naturally allows extension of control barrier functions to higher relative-degree systems, such as the payload-UAV system described in \ref{sec:sysModel}.

Suppose there exists an r-times continuously differentiable function $h(x):\mathcal{D} \rightarrow \mathbb{R}, \ \mathcal{D} \subset \mathbb{R}^n$, that is a function of the state vector $x \in \mathcal{D}$. Define a super level set $\mathcal{C}$ over $h$ as:

\begin{subequations}
\begin{align}
    \mathcal{C} &= \left\{ x \in D \subset \mathbb{R}^n : h(x) \geq 0\right\} \\
    \partial\mathcal{C} &= \left\{ x \in D \subset \mathbb{R}^n : h(x) = 0\right\} \\
    \text{Int}\left(\mathcal{C}\right) &= \left\{ x \in D \subset \mathbb{R}^n : h(x) > 0 \right\}
\end{align}
\label{eq:FISC}
\end{subequations}

where $\partial \mathcal{C}$ represents the boundary of the set $\mathcal{C}$ and $\text{Int}(\mathcal{C})$ represents the interior of the set $\mathcal{C}$. The set $\mathcal{C}$ is said to be \textit{forward invariant}, if for every $x(t_0) = x_0 \in \mathcal{C}$, the state trajectory $x(t) \in \mathcal{C}, \ \forall \ t > t_0$, i.e., the trajectory $x(t)$ never leaves $\mathcal{C}$ if it starts in $\mathcal{C}$. For a system whose dynamics can be written as $\dot{x} = f(x) + g(x)u$, the relative degree of $h(x)$ is defined as the number of times $h$ must be differentiated before the control input $u$ appears explicitly. Thus, the function $h$ is said to have a degree $r$ if $\mathfrak{L}_g\mathfrak{L}^{r-1}_fh(x) \neq 0$ and $\mathfrak{L}_g\mathfrak{L}_fh(x)$ = $\mathfrak{L}_g\mathfrak{L}^2_fh(x)$ = $\ldots$ = $\mathfrak{L}_g\mathfrak{L}^{r-2}_fh(x) = 0$, where $\mathfrak{L}_ab(x) = a \cdot \frac{\partial b}{\partial x}$ denotes the lie derivative of the vector field $b$ along the vector field $a$. \medskip

\begin{theorem} (\cite{nguyen2016exponential})
Given a set $\mathcal{C} \subset \mathbb{R}^n$ defined as a superlevel set of a r-times continuously differentiable function $h:\mathcal{D} \rightarrow \mathbb{R}$, then $h$ is an \textit{Exponential Control Barrier Function (ECBF)} if there exists a row vector $K \in \mathbb{R}^r$ such that
\begin{align}
    \inf_{u} \left[ \mathfrak{L}^r_fh(x) + \mathfrak{L}_g\mathfrak{L}^{r-1}_fh(x)u + K\eta(x) \right] \geq 0 \label{eq:cond_hx}
\end{align}
$\forall \ x \in \mathcal{C}$, where $\eta(x) = \left[ h(x) \ \dot{h}(x) \ \ldots \ h^{(r-1)}(x) \right]^T$. The set $\mathcal{C}$ is then a \textit{forward invariant} set.
\end{theorem}
\medskip

Consider the safe set $\mathcal{C} = \mathbb{R}^3 - \mathcal{O}$, where the obstacle set $\mathcal{O}$ is the space occupied by all the $N_{obs}$ obstacles. If it is ensured that the safe set $\mathcal{C}$ is \textit{forward invariant} for all time $t$, i.e., if the trajectory of the payload-UAV system starts in $\mathcal{C}$, and forever be trapped inside $\mathcal{C}$, then obstacle avoidance is guaranteed. To ensure that the set $\mathcal{C}$ is forward invariant, construct the $i^{th}$ Exponential Control Barrier Function $h_i(.):\mathbb{R}^3\rightarrow\mathbb{R}$ as:
\begin{align}
    h_i(x_c) = \left\Vert x_c - x_{obs_{i}} \right\Vert^2_{_2} - R_{o_i}^2, \ i = 1, \ldots, N_{obs}
    \label{eq:h_x}
\end{align}
where $x_{obs_i}, \ R_{o_i}$ are the position and the radius of the $i^{th}$ obstacle respectively, and $x_c$ is the position of a point $c_i$ on the surface of the convex hull around the payload-UAV system closest to the $i^{th}$ obstacle. It should be noted that the form of $h_i(.)$ is dependent on the $i^{th}$ obstacle shape. Eq. (\ref{eq:h_x}) holds for spherical obstacles only. However, Eq. (\ref{eq:h_x}) can be modified to include cylindrical pole obstacles (see Section \ref{subsec:gazebo}), by extending $z-$axis linearly in both positive and negative axis, and the vectors in the Eq. (\ref{eq:h_x}) only include $x-$, and $y-$ coordinates. It can be easily verified that $h_i(.)$ satisfies the properties described in Eq. (\ref{eq:FISC}). At this point, an assumption is made that the convex hull (and thus all the points on its surface) translates with a velocity equal to the translational velocity of the payload. This assumption is easily justified as long as the payload-UAV system remains close to the equilibrium point of Eq. (\ref{eq:EqPt}). From Eq. (\ref{eq:disSys}), it can be inferred that the dynamics for the payload is actually a double integrator system i.e., the control inputs appear only in the acceleration equations of the payload\footnote{One can substitute the A and B matrices given in the footnote \ref{ft:AB} in Eq. (\ref{eq:disSys}) to obtain the equations for the payload-UAV system.}. This results in the relative degree of $h_i(.)$ to be $2$. Thus, Eq. (\ref{eq:cond_hx}) reduces to:
\begin{align}
    \ddot{h}_i \geq -K \left[h_i \ \ \dot{h}_i\right]^T
    \label{eq:h_ddot}
\end{align}
where
\begin{subequations}
\begin{align}
    \begin{split}
    \dot{h}_i ={}& 2\left( x_c - x_{obs_i} \right)^T\left( \dot{x}_c - \dot{x}_{obs_i} \right) 
\end{split}\\
\begin{split}
    \ddot{h}_i ={}& 2\left( x_c - x_{obs_i} \right)^T\left( \ddot{x}_c - \ddot{x}_{obs_i} \right)\\
         & + 2\left\Vert \dot{x}_c - \dot{x}_{obs_i} \right\Vert_{_2}^2
\end{split}
\end{align}
\label{eq:OptCon}
\end{subequations}
and $K \in \mathbb{R}^2$ is a row vector, chosen such that the poles of the system (\ref{eq:h_ddot}) are all negative. Substituting acceleration equations of the payload for $\ddot{x}_c$ and substituting Eqs. (\ref{eq:OptCon}) in Eq. (\ref{eq:h_ddot}) and rearranging the equations results in a set of $N_{obs}$ constraints in $\Delta u_k$, of the form $\pmb{A}_{obs}\Delta u_k \leq B_{obs}$\footnote{The analytical forms of the matrix $\pmb{A}_{obs}$ and $B_{obs}$ can be found \href{https://drive.google.com/drive/folders/1ZIwZ4higpC-BZ5isK-Gr1uozPNqTscE7?usp=sharing}{here}. }.

Thus, the obstacle avoidance controller can be designed as follows:
\begin{subequations}
\begin{empheq}[box=\fbox]{align}
    {}& \Delta u_k^{*} = \argmin_{\Delta u_k} \frac{1}{2} \left\Vert\Delta u_k - \Delta \Bar{u}_k \right\Vert^2_{\pmb{Q}_{obs}} \\ \\
    \begin{split}\label{eq:con_avoid}
        {}& \text{s.t} \ \ \ \ \  \pmb{A}_{obs}\Delta u_k \leq B_{obs} \\
        {}& \left[\begin{array}{l} \pmb{I}_m \\ -\pmb{I}_m \end{array}\right]\Delta u_k \leq \left[ \begin{array}{c} \left( u_{ub} - u_e \right) \\ \left( u_e - u_{lb} \right) \end{array}\right]
    \end{split}
\end{empheq}
\label{eq:obsAvoid}
\end{subequations}
where $\pmb{Q}_{obs} > 0$ is a diagonal weighting matrix. The bounds on the control input are reapplied, along with the barrier function constraints. The optimal control law $\Delta u_k^*$ produced by the Integrated Decision Controller is thus equal to the optimal tracking control law $\Delta\bar{u}_k$ in the absence of obstacles, but while encountering the obstacles, the optimal control law $\Delta u_k^*$ is different, ensuring that the set of constraints in Eq. (\ref{eq:con_avoid}) is satisfied.

\section{Simulation Results}
\label{sec:simres}
In this section, the implementation details and simulation results are discussed to evaluate the performance of the proposed controller. The Linear Model Predictive Controller is implemented in \texttt{C++} using the \texttt{ACADO} toolkit \cite{Houska2011a} code generator. \textit{Multiple Shooting} discretization technique is employed to obtain a discrete system analogous to Eq. (\ref{eq:disSys}). The prediction horizon for the LMPC is chosen to be $20$ time steps. The optimization problems in both Eq. (\ref{eq:JMPC}) and  Eq. (\ref{eq:obsAvoid}) is solved using the \textit{online active set strategy} \cite{Ferreau2008} as implemented in the \texttt{qpOASES} library \cite{Ferreau2014}. To construct the convex hull, and find the point on the surface of the convex hull closest to the obstacle, \textit{polytope distance} algorithms \cite{cgal:hs-ch3-21b} are used as implemented in the \texttt{CGAL} library \cite{cgal:eb-21b}.

The advantage of detaching the obstacle avoidance controller from the MPC is that it enables the user to use pre-existing, state-of-art MPC implementations that are highly optimized for nonlinear systems, and semi-definite hessian matrices that occur in the MPC optimization stage, as discussed in \cite{Houska2011a}.

\begin{table}
    \centering
    \caption{Parameter values for the payload-UAV system }
    \label{tab:paramVals}
    \begin{tabular}{ccccc} \toprule
         & $m$ & $J_{xx}$ & $J_{yy}$ & $J_{zz}$ \\ \midrule
         \text{Payload} & 3.1 & 0.29 & 0.29 & 0.55 \\
         \text{UAVs} & 0.7 & 0.01 & 0.01 & 0.01 \\ \midrule
         & & $p_i$ & & $l_i$ \\ \midrule
         \text{UAV1} & & $[0.5, \ 0.5, \ \shortminus0.25]^T$ & & 3.2 \\
         \text{UAV2} & & $[0.5, \ \shortminus0.5, \ \shortminus0.25]^T$ & & 3.2 \\
         \text{UAV3} & & $[\shortminus0.5, \ \shortminus0.5, \ \shortminus0.25]^T$ & & 3.2 \\
         \text{UAV4} & & $[\shortminus0.5, \ 0.5, \ \shortminus0.25]^T$ & & 3.2 \\\bottomrule
    \end{tabular}
\end{table}

\subsection{Numerical Results}
For the subsequent discussions, the parameters considered for the payload-UAV system is shown in Table \ref{tab:paramVals}, and the constants $\delta_{\phi}, \delta_{\theta}, \delta_{\psi}$ of Eq. (\ref{eq:consOsc}) are set to $5$ degrees. It is observed that assigning a value lower than 5 degrees to these constants renders the optimization problem of Eq. (\ref{eq:J_MPC}) infeasible, especially during the instants where obstacle avoidance occurs.
\subsubsection{Trajectory Tracking}
\begin{figure}
    \includegraphics[scale=0.65]{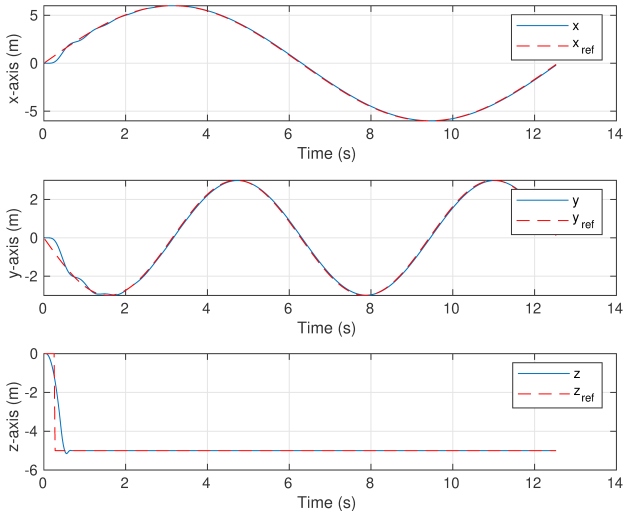}
    \caption{Position of the payload is shown in solid line, and the desired trajectory is shown in dashed lines (no obstacles)}
    \label{fig:trackxyz}
\end{figure}

\begin{figure}
    \includegraphics[scale=0.65]{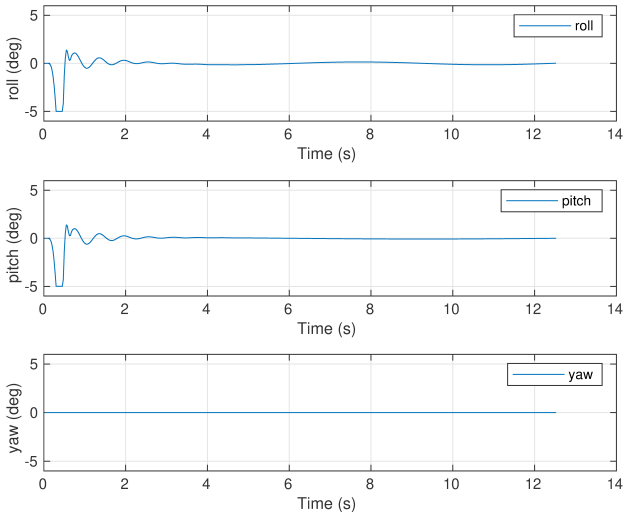}
    \caption{Attitude of the payload (no obstacles)}
    \label{fig:attrpy}
\end{figure}

\begin{figure}
    \includegraphics[scale=0.45]{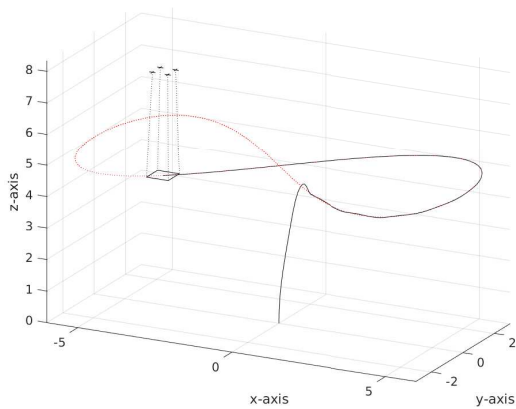}
    \caption{A snapshot of the numerical simulation. The reference trajectory is shown in red dotted lines, and the payload trajectory is shown in black line. For the purpose of demonstration, the coordinates are converted to ENU (east, north, up) from the NED coordinates.}
    \label{fig:anim3d}
\end{figure}

For a numerical evaluation of the algorithm, four UAVs are considered that transport a rigid payload, as shown in Fig. \ref{fig:schDiag}. The desired trajectory for the payload is chosen as a figure-eight $\infty$ contour, as described by Eq. (\ref{eq:Lissajous}):
\begin{subequations}
\begin{align}
    r_{{0d}_x} &= 6\sin\left( 0.5t \right) \\
    r_{{0d}_y} &= -6\sin \left( 0.5t \right) \cos \left( 0.5t \right) \\
    r_{{0d}_z} &= -5u(t)
\end{align}
\label{eq:Lissajous}
\end{subequations}

where $u(t)$ is the unit step function. From Fig. \ref{fig:trackxyz} and \ref{fig:attrpy}, it can be seen that the LMPC controller successfully stabilizes the payload, while ensuring that it tracks the desired trajectory. The attitude of the payload is limited to within $5$ degrees during the abrupt take off, but dies down quickly once stabilized. The yaw angles of the payload and UAVs are highly penalized, to ensure that the optimization problem remains strictly convex\cite{wehbeh2020distributed}. A higher cost is placed on the attitude and the position of the payload, and a slightly lower cost is placed on the attitude of the links and UAVs in the weighting matrix $\pmb{Q}$ of Eq. (\ref{eq:JMPC}). This ensures that the system can still make slight deviations from its equilibrium points, making the system less stiff, and slightly more flexible. A snapshot of the 3D simulation\footnote{\label{ft:video} The video demonstrations for both the numerical simulations as well as \texttt{Gazebo} simulations can be found here:  \href{https://youtu.be/AMVlMNYQCLw}{https://youtu.be/AMVlMNYQCLw} } is shown in Fig. \ref{fig:anim3d}, where the complete scenario is shown. 

\begin{figure}
    \centering
    \includegraphics[scale=0.65]{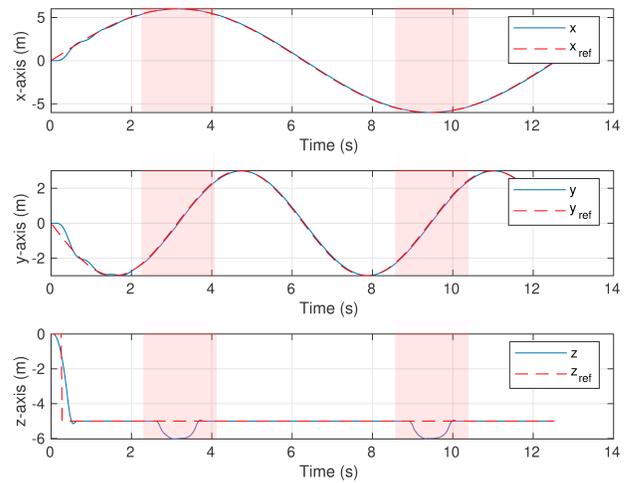}
    \caption{Trajectory tracking with two surrounding obstacles. The instants of obstacle avoidance is shown in red shaded area.}
    \label{fig:trackObsxyz}
\end{figure}

\begin{figure}
    \centering
    \includegraphics[scale=0.65]{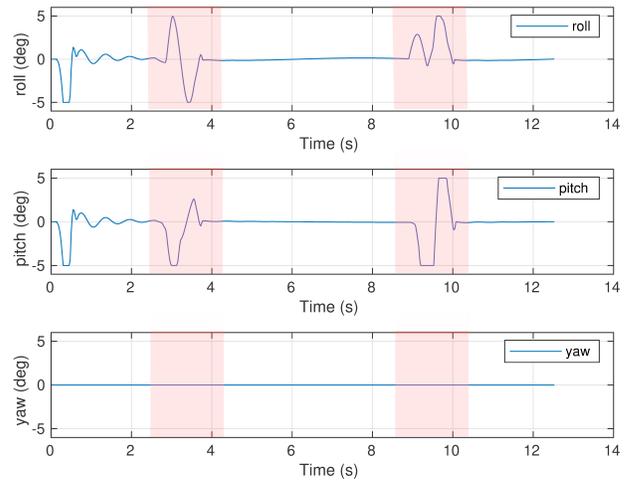}
    \caption{Attitude of the payload with two surrounding obstacles. The instants of obstacle avoidance is shown in red shaded area.}
    \label{fig:attObsrpy}
\end{figure}

\begin{figure}
    \centering
    \includegraphics[width=7cm, height=6cm]{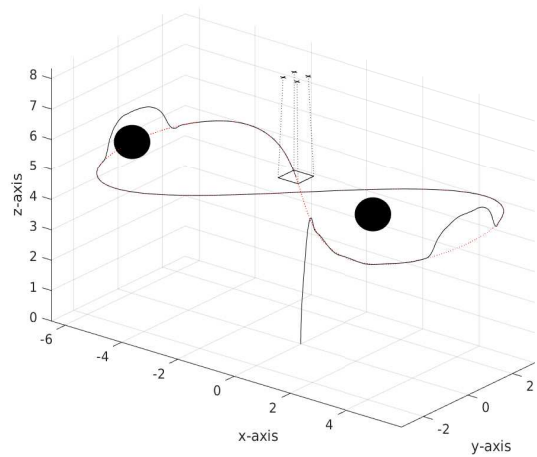}
    \caption{A snapshot of the obstacle avoidance simulation. The sphere on the left is a static obstacle and the sphere on the right is a dynamic obstacle.}
    \label{fig:animObs3d}
\end{figure}

\begin{figure*}
\centering
\includegraphics[scale=0.75]{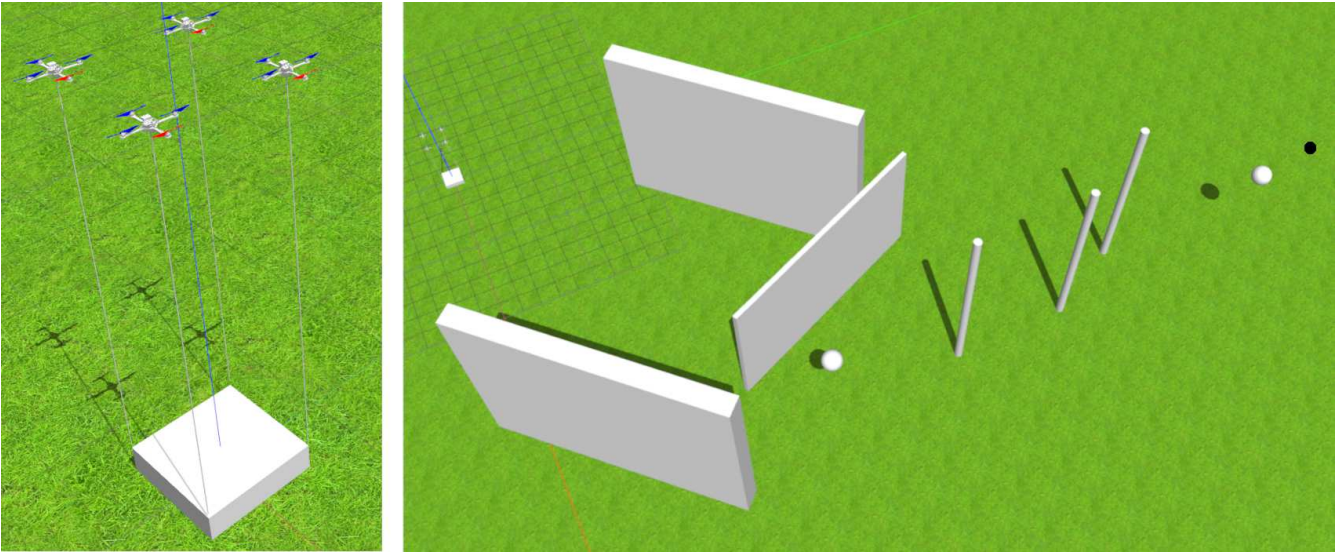}
\caption{Figure on the left shows the payload-UAV system, and the figure on the right shows the obstacle-avoidance environment course, that consists of walls, poles and spherical obstacles. The goal point for the payload is shown by a small black circle behind the rightmost sphere.}
\label{fig:gazObs}
\end{figure*}

\begin{figure}
    \centering
    \includegraphics[scale=0.65]{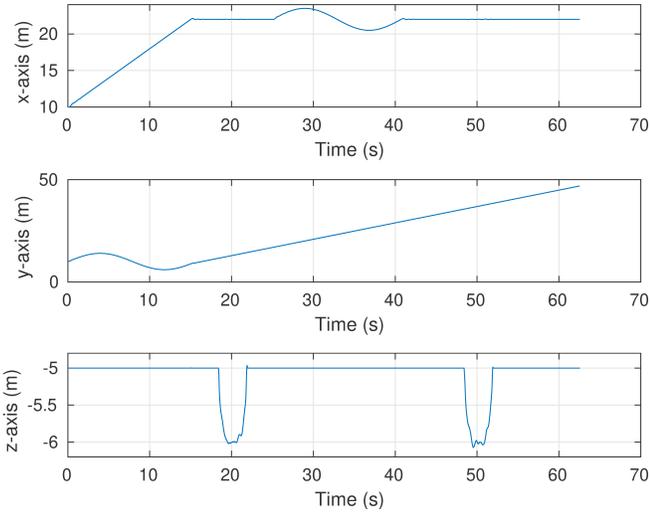}
    \caption{Payload Trajectory for the obstacle environment course in \texttt{Gazebo}.}
    \label{fig:payloadObsGaz}
\end{figure}

\begin{figure}
    \centering
    \includegraphics[scale=0.225]{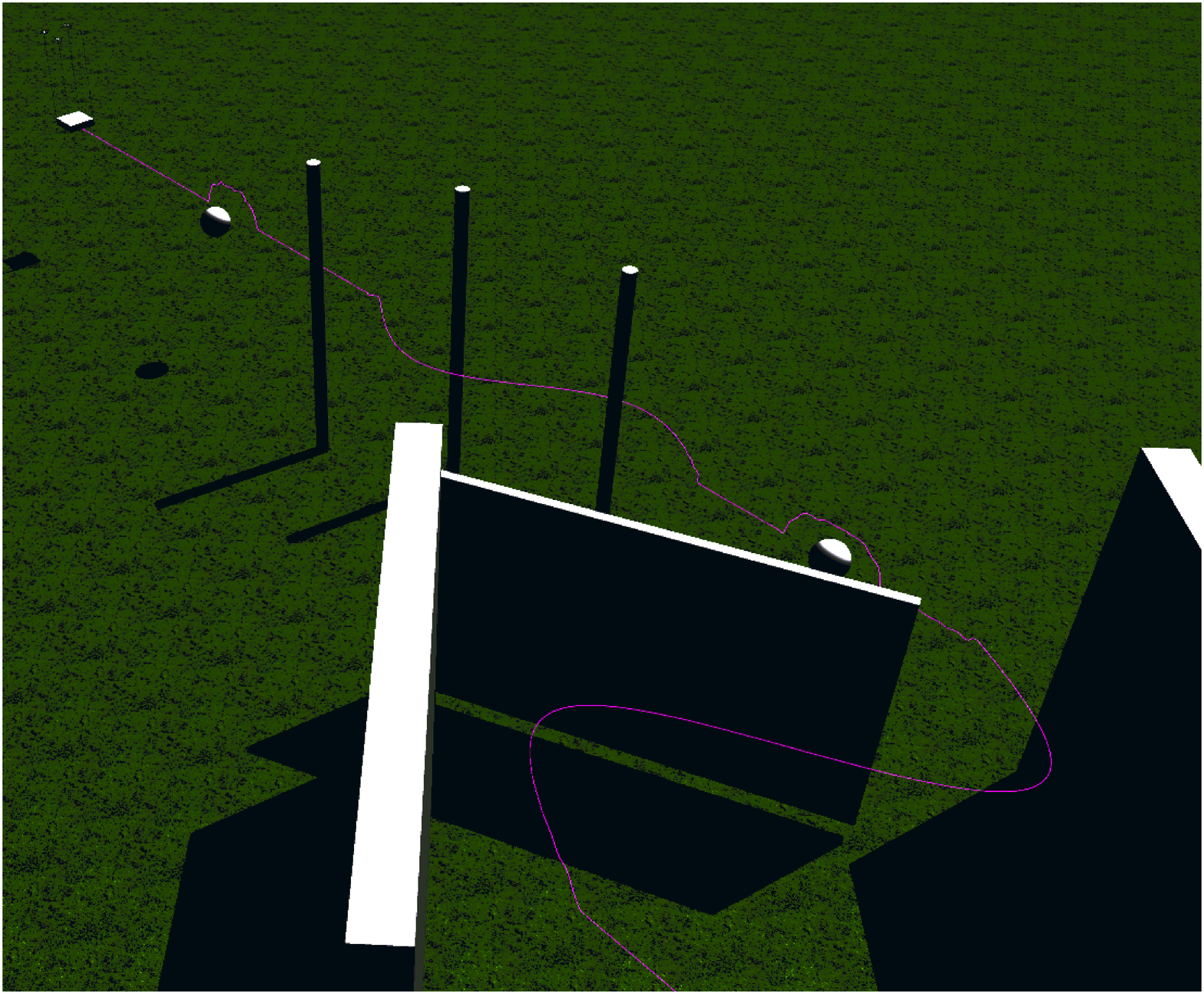}
    \caption{The purple line shows the trajectory of the payload. The image is darkened to make the purple line more prominent and noticeable.}
    \label{fig:payTrajGaz}
\end{figure}

\subsubsection{Obstacle Avoidance}

The numerical simulation consists of both static and dynamic obstacles. It is assumed that the obstacles can be enclosed in sphere of radius $R_o$. For demonstration purposes, the obstacle radius is considered to be $0.5$m and a safety margin of $0.5$m around the obstacle is taken. Thus, $R_0 = 1$ meter. The dynamic obstacle is modelled as a simple harmonic oscillator, oscillating about its mean position $(6, 0, \shortminus5)$m. The mean position of the dynamic obstacle coincides with a point on the desired trajectory, and the obstacle timing is set in such a way that it directly confronts the payload-UAV system at its mean position. The static obstacle is placed at $(\shortminus6, 0, \shortminus5)$m. Due to the dynamic nature of the obstacle, the constraint matrices $\left(\pmb{A}_{obs}, B_{obs}\right)$ of Eq. (\ref{eq:obsAvoid}b) are time-varying in nature. 

It can be observed from Fig. \ref{fig:trackObsxyz} that the trajectory only changes along the $z-$axis, where the payload goes above the obstacle, as this is the most feasible thing to do. As seen in Fig. \ref{fig:attObsrpy}, the payload oscillates while avoiding the obstacles. This oscillation is however bounded under a small deviation of $5$ degrees and occurs due to the fact that the payload-UAV system is avoiding the obstacle while in motion, without slowing down. Due to the high cost placed on the yaw angles of the payload and UAVs, there is almost no change in the yaw configuration of the system throughout the course of motion. A bird's eye snapshot of the 3D obstacle avoidance simulation\footnoteref{ft:video} is shown in Fig. \ref{fig:animObs3d}. The spherical obstacle on the right oscillates along the $x-$axis, and periodically meets the payload-UAV system at the point $(6, 0, \shortminus5)$, while the spherical obstacle on the left remains stationary at the point $(\shortminus6, 0, \shortminus5)$.

\begin{figure}
    \centering
    \includegraphics[scale=0.55]{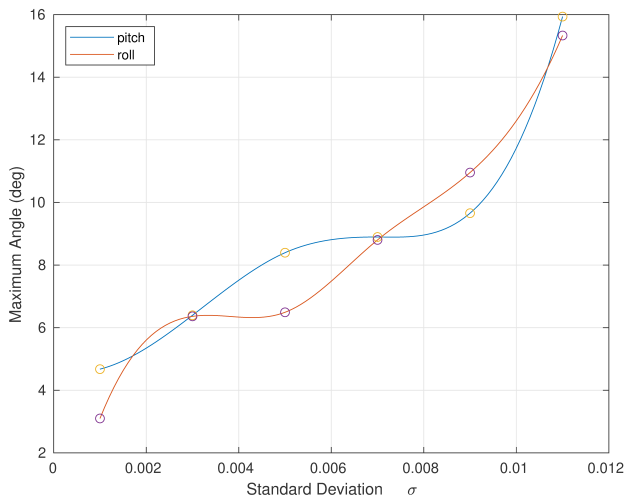}
    \caption{Maximum roll and pitch angle vs the noise standard deviation $\sigma$ ($m/s$ for linear velocity) and ($rad/s$ for angular velocity}
    \label{fig:sigmavsdev}
\end{figure}

\subsection{Gazebo Simulation}
\label{subsec:gazebo}
The obstacle avoidance control algorithm is validated by testing it inside a high-fidelity simulation environment. The \texttt{Gazebo} simulator \cite{Koenig-2004-394} is used to create an obstacle-avoidance environment course. For spawning the payload-UAV system, the entire model is written in a \texttt{urdf} file, and the \texttt{RotorS} package \cite{Furrer2016} is used for spawning four hummingbird drones. This is shown in Fig. \ref{fig:gazObs}. The control algorithm is implemented in a \texttt{C++} script, which communicates with the \texttt{Gazebo} simulator via the \texttt{ROS} library. The inputs to the model in the simulation are rotor rpm values, which can be easily obtained from the thrust force and torques via a linear transformation matrix.

The environment in \texttt{Gazebo} consists of many obstacles including walls and poles that are static and spherical obstacles that are dynamic. The desired trajectory that the payload must track goes in between the walls, and then to the goal point behind the right-most sphere (see Fig. \ref{fig:gazObs}). This ensures that the payload-UAV system doesn't go all the way around the walls and then to the goal point. 

The position of the leftmost and rightmost spherical obstacle is at $(22, 13, 5)$ and $(22, 37, 5)$ respectively. The cylindrical poles are placed at $(22, 20, 5)$, $(22, 26, 5)$, $(22, 30, 5)$ from left to right. It must be noted that the coordinate conventions in \texttt{Gazebo} simulator is ENU (East-North-Up). Because of this, there is a coordinate conversion between NED to ENU at the interface of the \texttt{ROS C++} script and the \texttt{Gazebo} APIs.

It can be observed from Fig. \ref{fig:payloadObsGaz} that the payload successfully avoids both the spherical obstacles, as well as the cylindrical pole obstacles. Unlike for spherical obstacles where the payload dodges by going above them, due to the height of the pole obstacles, the payload goes around them by maintaining a safe distance from these poles. This is inferred better\footnoteref{ft:video} from Fig. \ref{fig:payTrajGaz}, where the purple line shows the trajectory history of the payload. The sphere near the walls is a static sphere, where as the sphere towards the end is a dynamic sphere oscillating about its mean position $(22, 37, 5)$. The control algorithm is able to run at a real-time frequency of 20Hz, providing optimal rpm values to the UAVs in the simulation.

\subsection{Ablation Study}
To evaluate the robustness of the proposed IDC, some of the practical uncertainties are explored in the following paragraphs.
\subsubsection{Mass Uncertainty}
In practice, there is some uncertainty between the reported mass of the payload and the actual mass of the payload. A $\pm 10\%$ variation is considered here. From Table \ref{tab:paramVals}, the value for the payload mass is taken as $3kg$ during controller design, but the payload mass in the plant equations is chosen to be either $2.7kg$ or $3.3kg$. It is observed that the performance of the controller doesn't change, however the control effort $\left\Vert u \right\Vert_2$ required varies by $\pm 5\%$. This is justified, as the UAVs need to compensate for the change in the weight, while meeting the tracking requirements.

\subsubsection{Noisy Payload States}
The linear velocity and angular rates of the payload in Eq. (\ref{eq:stateqn}) is subjected to zero mean Gaussian noise $\mathcal{N}(0, \sigma)$. This is done to mimic the actual payload state values obtained from a noisy rate sensor. For now, the UAV states are ignored, due to the presence of onboard estimation algorithms that can accurately infer the UAV states. The goal is to analyze the robustness of the proposed IDC to the noisy state values of the payload i.e., to find the maximum value of the standard deviation $\sigma$ ($m/s$ for linear velocity and $rad/s$ for angular velocity), such that the payload oscillations remain under a certain threshold. The results are shown in Fig. \ref{fig:sigmavsdev}. It can be observed that the maximum roll and pitch angles of the payload increase monotonically as the standard deviation $\sigma$ increases. Beyond a standard deviation of $0.012$, the obstacle avoidance controller becomes unstable, while the tracking controller continues to track the trajectory (with a noisy performance) when assessed independently. If the maximum threshold for payload oscillations is set at $10$ degrees, the standard deviation must not be above $0.008$. In practice, most of the IMU's standard deviation falls below or occurs close the nominal value of $0.008$\footnote{A list of parameters for a few commonly used rate sensors are available here: \href{https://github.com/rpng/kalibr_allan}{https://github.com/rpng/kalibr\_allan}}. It is also observed that the maximum roll and pitch angles occur only at the time of avoiding the obstacles.

\subsubsection{Relative Safety Margin}
The IDC ensures that there is minimum deviation from the reference trajectory while avoiding the obstacles. Earlier, a safety margin of $0.5$m was considered. It is observed that by changing the safety margin the controller performance doesn't change; it only determines the relative proximity between the payload-UAV system and the obstacle. In practice, the relative distance to the obstacles may be underestimated or overestimated by some value. Thus, considering the right safety margin is crucial. Setting a higher safety margin may result in a higher deviation from the reference trajectory, which in some cases may render the tracking controller unstable whereas setting a lower safety margin may be undesirable, as the payload-UAV may come too close to the obstacle, which could result in a collision. After various experimentation, a thumb rule of choosing the safety margin equal to the obstacle radius is found to provide best results.

\section{Conclusions}
\label{sec:conclusion}
In this paper, an integrated decision control-based obstacle avoidance controller is presented to solve the problem of collaborative payload transportation in a cluttered environment. The IDC fuses the optimal tracking control provided by the Model Predictive Controller, with the safety-critical constraints provided by the Exponential Control Barrier Functions in an optimization framework. The structure of the ECBFs depend on the shape of the obstacle, which can be static or dynamic in nature. To ensure that the UAVs or the payload don't collide with the surrounding obstacles, a safe convex hull is computed around the entire system at each time instant, and the point on the convex hull closest to the surrounding obstacle is used to generate the ECBF constraints. Numerical simulations are conducted on the proposed controller to demonstrate its functionality, and eventually the algorithm is deployed in a real time, high fidelity simulation using \texttt{Gazebo}. An ablation study is conducted to demonstrate the robustness of the IDC to handle practical unforeseen situations like payload mass uncertainties, noisy payload states and choosing an optimal relative safety margin around the obstacles.

In the scope of future work, non-linear control laws will be explored that can handle the problem of tracking and obstacle avoidance in a single framework. Moreover, external disturbances and state estimation techniques can be used to make the controller more robust to sensor noise and model uncertainties. Further, with the help of additional supporting mechanisms like rack-pinion joints, formation control of the UAVs for rigid payload can be made possible.

\section{Acknowledgements}
The authors would like to acknowledge the financial support from the Nokia CSR grant on Network Robotics.

\bibliographystyle{IEEEtran}
\bibliography{references}
\nocite{*}

\end{document}